\documentclass[a4paper,10pt]{article}

\usepackage{fourier}
\usepackage{color}
\usepackage{graphicx}
\usepackage{url}
\usepackage[affil-it]{authblk}
\usepackage{amsmath}
\usepackage{wrapfig}
\usepackage{xspace}

\usepackage[T1]{fontenc}
\usepackage{times}

\usepackage{times}
\usepackage{epsfig}
\usepackage{graphicx}
\usepackage{amsmath}
\usepackage{amssymb}
\usepackage{color}
\usepackage{tabularx, booktabs}
\usepackage{hyperref}
\usepackage{wrapfig}

\begin{document}

\title{Fast and Accurate Optical Flow based \\ Depth Map Estimation from Light Fields}
\date{\vspace{-5ex}}

% % Enter the paper's authors in order
% \author{}
\author{Yang Chen}%{cheny5@scss.tcd.ie}{1}
\author{Martin Alain}%{alainm@scss.tcd.ie}{1}
\author{Aljosa Smolic}%{smolica@scss.tcd.ie}{1}

% % Enter the institutions
% % \addinstitution{Name\\Address}
\affil{
 V-SENSE project\\
 Graphics Vision and Visualisation group (GV2)\\
 Trinity College Dublin\\
 %Dublin, Ireland
}

\maketitle
\thispagestyle{empty}

\begin{abstract}

Depth map estimation is a crucial task in computer vision, and new approaches have recently emerged taking advantage of light fields, as this new imaging modality captures much more information about the angular direction of light rays compared to common approaches based on stereoscopic images or multi-view.
In this paper, we propose a novel depth estimation method from light fields based on existing optical flow estimation methods.
The optical flow estimator is applied on a sequence of images taken along an angular dimension of the light field, which produces several disparity map estimates.
Considering both accuracy and efficiency, we choose the feature flow method as our optical flow estimator.
Thanks to its spatio-temporal edge-aware filtering properties, the different disparity map estimates that we obtain are very consistent, which allows a fast and simple aggregation step to create a single disparity map, which can then converted into a depth map.
Since the disparity map estimates are consistent, we can also create a depth map from each disparity estimate, and then aggregate the different depth maps in the 3D space to create a single dense depth map.

\end{abstract}

%-------------------------------------------------------------------------
\section{Introduction}
\label{sec:intro}

Light fields aim to capture all light rays passing through a given volume of space~\cite{Levoy1996}.
Compared to traditional 2D imaging systems which capture the spatial intensity of light rays, a 4D light field contains the angular direction of the rays.
Light fields have thus become a topic of growing interest in several research areas such as image processing, computer vision, and computer graphics.
Applications include refocusing of an image after capture, rendering new images from virtual points of view, or computational displays for virtual and augmented reality.
In this paper, we focus on depth map estimation from a light field.

3D scene reconstruction or depth estimation from light fields is a major topic of interest and many methods have been proposed in the past years.
Several methods have been proposed which estimate disparity between views of the light field with respect to the center view using existing stereo-matching techniques~\cite{Jeon2015,Neri2015}.
To better exploit the light field structure, novel approaches have been introduced relying either on angular patch analysis~\cite{Chen2014b,Wang2015a,Si2017} or the Epipolar Plane Images (EPI) representation of light fields~\cite{Wanner2012a,Zhang2016a,Kim2013,Johannsen,Tao2013}.

%Conventonal methods are mainly based on stereo-matching, which may gain local accuracy, but lost coherent information globally.

In this paper, we present a novel pipeline to estimate depth maps from light fields based on optical flow, where the measured displacements correspond to disparity, from which we can then obtain the depth.
%using coherent light field spatial-angular information with optical flow methods.
Our main contributions are:
(a): We propose a novel depth map estimation scheme based on an spatio-angular edge aware optical flow~\cite{lang2012practical} applied over an angular dimension of the light field.
%efficient temporal(angular) filtering by using a edge aware filtering, 
%which is tailored to the structure of 4D light field data and robust than conventional stereo-matching methods.
(b): We propose to combine the spatio-angular optical flow with the state-of-the-art coarse-to-fine patch matching method~\cite{hu2016efficient} as initialization, which significantly improves the results without increasing the running time.
(c): We propose to directly combine our multiple and consistent depth map estimates in the 3D space to obtain very dense depth maps or point clouds.
% (c): We show the outperformed balance of accuracy and efficiency of our scheme by achieving competitive accurate results of state-of-art methods with significant speed-up.
We show that the proposed approach achieves comparable performances to the best state-of-the-art method in terms of balance between speed and accuracy.

This paper is organized as follows.
In section \ref{sec:relatedWork} we review the 4D structure of light fields and existing methods to retrieve depth from light fields, as well as the state-of-the-art optical flow estimation techniques.
Section \ref{sec:approach} describes more in details the proposed approach.
Finally, in section \ref{sec:eva} we evaluate the performance of our method.

\section{Background and related work}
\label{sec:relatedWork}

\subsection{Light field 4D structure}

\begin{figure}[t!]
    \begin{center}
    \includegraphics[width=0.9\linewidth,trim={0cm 0cm 0cm 7cm},clip]{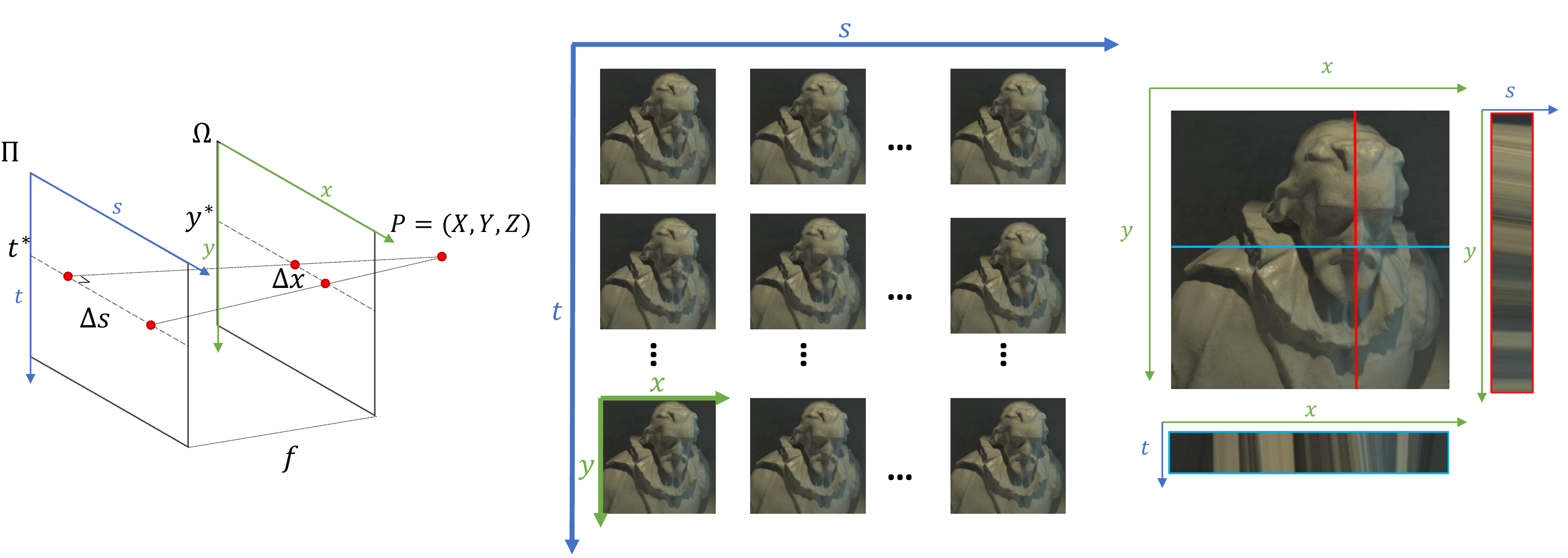}
    \end{center}
    \caption{Light field two-plane parametrization (left), matrix of views representation (middle), and Epipolar Plane Images (EPI) representation (right).}
    \label{fig:light_field_geo}
    \vspace{-0.5cm}
\end{figure}

%Light fields sample all light rays going through a given volume of space, and capture in addition to the light intensity the angular direction of each light ray.
We adopt in this paper the common two-plane parametrization as shown in Figure~\ref{fig:light_field_geo}, and a light field can be formally represented as a 4D function $\Omega \times \Pi \to \mathbb{R}, (x, y, s, t) \to L(x, y, s, t)$ in which the plane $\Omega$ represents the spatial distribution of light rays, indexed by $(x, y)$, while $\Pi$ corresponds to their angular distribution, indexed by $(s, t)$.

Perhaps the easiest way to visualize a light field is to consider it as a collection of views, also called sub-aperture images, taken from several view points parallel to a common plane.
The light field can then be considered as a matrix of views (see Figure~\ref{fig:light_field_geo}).
Note that an important assumption when using such representation is that the different views are rectified.
%In this paper, each angular direction is used as the same way as temporal direction in common image sequence consideration.
Another common representation of light fields are Epipolar Plane Images (EPI), which are 2D slices of the 4D light field obtained by fixing one spatial and one angular dimension ($sy$- or $xt$-planes, see Figure~\ref{fig:light_field_geo}).

Note that light fields can be captured using lenslet camera such as Lytro \cite{Lytro} or camera arrays, however we use in our experiments synthetic light fields for which the depth map ground truth is known.

\subsection{Depth estimation from light fields}
\label{DPfromLF}

A 4D light field implicitly captures the 3D geometry of a scene.
As illustrated in Figure~\ref{fig:light_field_geo}, the depth $Z$ of a point $P$ in 3D space can be obtained as: $Z = -f\frac{\Delta s}{\Delta x} \label{eq:depth_vs_disp}$, where $f$ and $\Delta s$ are respectively the focal length and the distance between camera positions, which are known parameters at the time of capture.
The depth $Z$ can thus be obtained by estimating the disparity $\Delta x$.

Multiple methods taking advantage of the existing literature in stereo disparity estimation have then been proposed to estimate depth from light fields.
These techniques rely on various matching approaches to estimate the disparity between views from the light field and a reference view (often the center one).
In~\cite{Jeon2015}, the authors proposed an accurate block-matching method reaching sub-pixel accuracy based on the Fourier phase-shift theorem.
%In \cite{Sabater2015}, the authors proposed a complete pipeline to decode views from light field captured with a lenslet camera and then estimate the disparity based on robust block-matching \cite{Sabater2011}.
%In \cite{Navarro2017}, sparse and accurate matchings are first found, and then interpolated using optical flow.
%To obtain precise matching at a sub-pixel accuracy, a method based on Fourier phase-shift theorem was proposed in \cite{Jeon2015}.
To reduce the complexity, a multi-resolution approach was proposed in~\cite{Neri2015}.
%In \cite{Heber2014a}, a global matching is obtained through a low-rank decomposition of the views, refined with homographies.

To better take into account the light field structure, extensions of the previous methods have been proposed based on the analysis of texture patches sampled along the angular dimensions instead of the spatial dimensions.
These angular patches, also called SCam, were first exploited in~\cite{Chen2014b}.
This work was further extended in~\cite{Wang2015a} to be robust to occlusion.
More recently, this idea was included in a global optimization framework~\cite{Si2017} in order to obtain a dense depth map estimation.

% The epipolar lines are also used in \cite{Jeon2015} to correct distortions before applying stereo-matching techniques on the views of the light fields (corresponding to the 2D spatial slices over the $xy$ coordinates of the 4D light field, see Fig. \ref{fig:light_field_geo}).
%Sub-pixel accuracy of the matchings is reached using the Fourier phase-shift theorem.

Several techniques also exploit the light field structure through EPI, as in such images the slope of a line has a linear relationship with the depth.
In~\cite{Wanner2012a}, the slope of the epipolar lines are estimated using a structure tensor, while in~\cite{Zhang2016a} a spinning parallelogram operator is proposed.
In~\cite{Kim2013}, depth from high spatio-angular resolution light fields is obtained by first estimating high confidence depth values on the EPI edges, and then propagating this information to homogeneous regions using a fine-to-coarse approach.
In~\cite{Johannsen}, a sparse decomposition of the EPI is performed over a depth-based dictionary built from fixed disparities, and the scene disparity is deduced from the sparse coding coefficients.
In~\cite{Tao2013}, defocus and correspondence cues are obtained from the EPI by shearing the epipolar lines.
The shearing angle optimizing the multiple cues response gives the slope of the epipolar lines, and thus the depth.

Note that most of the aforementioned methods require an additional regularization or optimization step, which is usually a computationally intensive global process.
%Popular methods include graph cuts \cite{Kolmogorov2002}, coarse-to-fine approaches \cite{Graber2015}, or total generalized variation \cite{Bredies2010}.

\subsection{Optical flow}

Since Horn and Schunck's pioneer work~\cite{HS} in variational optical flow estimation, many methods have been proposed exploring their idea based on energy minimization.
%many extended applications has been widely explored in the energy optimization fashion.
In this subsection, we will briefly introduce the related literature in this field.
For a comprehensive survey, please refer to the work in~\cite{Baker2011,li2014robust}. %,li2013optical,barron1992performance,li2013nonrigid,li2017blur}.

The original work from~\cite{HS} often leads to inaccurate estimation of large pixel displacements. % of the real-world images.
To improve the accuracy of such challenging cases, the patch match method~\cite{barnes2009patchmatch}, initially introduced for nearest neighbor field estimation, has been adapted to the optical flow problem.
%Given this background, Patch match~\cite{barnes2009patchmatch} is proposed to approximate the nearest neighbor field(NNF) which represents the sparse pixel correspondence.
Patch match provides sparse pixel correspondence, and the final optical flow estimation is then considered as a labeling problem leveraging the coherent information of natural images.
To further obtain a dense flow map, Revaud~et al.~\cite{revaud2015epicflow} propose an edge-preserving interpolation scheme applied on top of sparse matching correspondences. % which is reported efficient if this is extended to hierarchical fashion~\cite{hur2015generalized}.
In this context, Hu~et al.~\cite{hu2016efficient} propose a coarse-to-fine extension of the basic patch match method, which is proven efficient in finding reliable correspondences on large pixel displacements.
%Many further spatial based approaches~\cite{li2014robust,li2017nonrigid} give improvement on noisy and occlusion handling.

%However, like most usual optical flow method, these method still only focus on spatial relationship in pairwise image although there is more coherent information waits to be explored. Besides, many past work also presented various way to transfer sparse matches to dense optical flow, such as~\cite{revaud2015epicflow} proposed an edge-perserving interpolation scheme based on sparse matching correspondence,~\cite{hur2015generalized} explored this with the hierarchical matching, and~\cite{brox2011large} modified classic variational minimization equation by adding an extra matching term.

In addition to the spatial accuracy, the optical flow temporal consistency has been an important and challenging research topic.
%has been considered as an important role in addition to the spatial constraints.
In their early work, Murray~et al.~\cite{murray1987scene}, proposed to add a temporal smoothness term to improve the temporal consistency.
Sliding windows~\cite{volz2011modeling} and Kalman filtering~\cite{hoeffken2011temporal} based methods have been proposed later, focusing on temporal stability, although their performances highly rely on the selection of the window size.
Feature flow~\cite{lang2012practical} proposed a novel local edge-aware filtering to replace the expensive global optimization used in previous work, which significantly reduced the computation cost while performing an accurate estimation.
%Meanwhile, feature flow also has no worry about window size problem due to it process entire video at each running time.
However, this method relies on a sparse correspondence initialization, which has a significant impact on the final result.

%In this paper, we use the accurate CPM as initialization of the feature flow to obtain a fast and accurate optical flow estimation method.
% we propose a rapid and accurate method which first estimates the disparity, then obtain the depth map via basic equation.
% However, contrary to the methods described above, we do not rely on stereo-matching techniques, but instead use temporally consistent optical flow method with state-of-the-art patch match as initialization, applied on a 3D spatio-angular volume extracted from the light field.
% In this way, our method does not require a computationally intensive optimization step.

\section{Depth estimation from light fields using optical flow}
\label{sec:approach}

\subsection{System overview}

\begin{figure}[t]
	\begin{center}
	\includegraphics[width=0.9\linewidth,trim={0cm 5.5cm 0cm 0cm},clip]{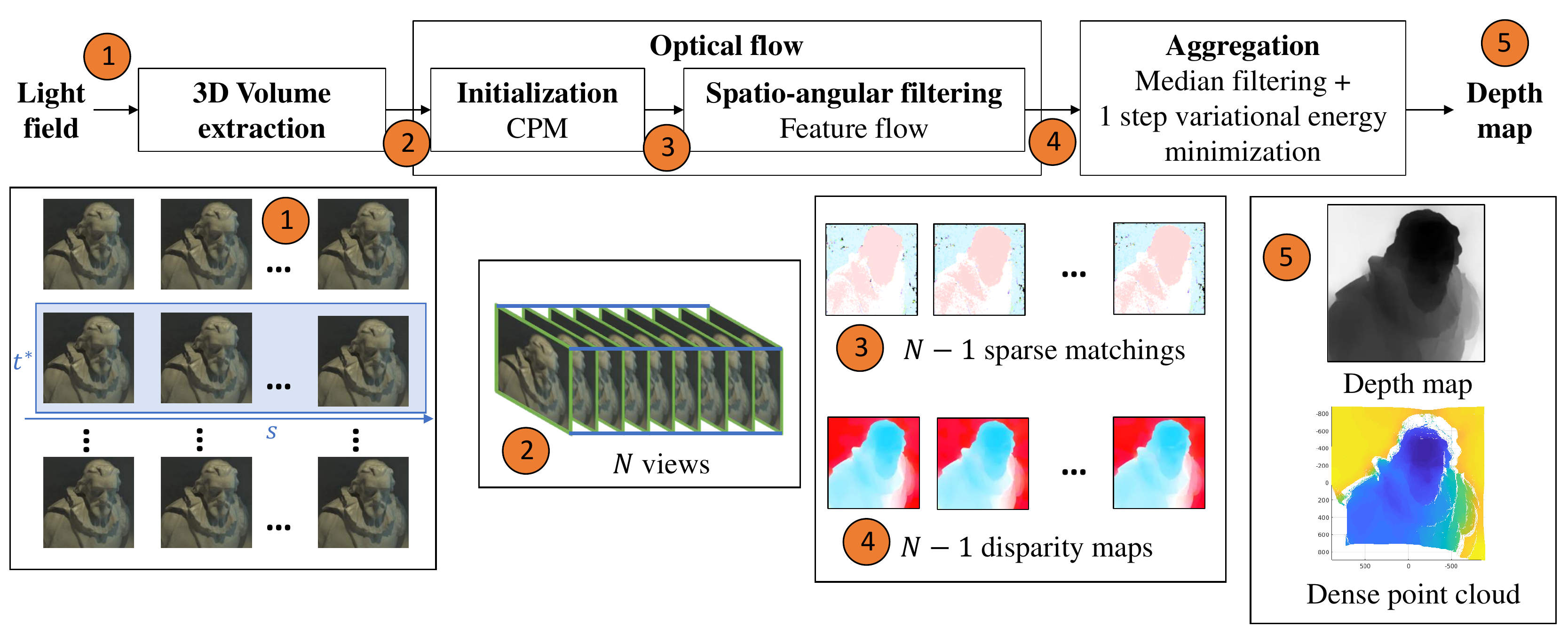}
	\end{center}
	\vspace{-0.5cm} 
	\caption{Overview of the proposed approach. The method can be apply on any number of rows or columns of the matrix of views in order to obtain more disparity map estimates.}
\label{fig:pipeline}
\vspace{-0.2cm}
\end{figure}

In this paper, we propose a novel scheme for efficient and accurate estimation of depth maps from the 4D structure of light fields using optical flow.
Our approach consists of three main steps, as illustrated in Figure~\ref{fig:pipeline}.
First, a 3D spatio-angular volume is extracted from the light field by taking views along a given angular dimension.
%This sequence is then used as an input for an optical flow estimation.
Second, an optical flow estimation is performed over this spatio-angular volume.
The displacements measured by the optical flow thus correspond to disparity estimates between consecutive views of the light field.
%Note that 
The optical flow itself consists of two steps:
an initialization with a sparse matching correspondence technique~\cite{hu2016efficient} is first performed, then a spatio-angular edge-aware filter is applied on the sparse estimates to obtain a dense flow estimation.
For this volumetric filtering we choose the feature flow method~\cite{lang2012practical}.
Finally, an aggregation step is performed to obtain the depth map from the multiple disparity map estimates.
A common process used in most state-of-the-art methods performs sophisticated weighted median or image guided filtering combined with costly global energy minimization on the disparity map estimates to obtain a single accurate disparity map. Thanks to the the edge-aware filtering along the angular dimension of the optical flow, which enforces the consistency between the different disparity map estimates, we can reduce this step to a simple median filtering followed by a one-step variational energy minimization.
For the same reason, we can propose in this paper a novel process where several depth maps are created from each disparity map estimate, and then fused in the 3D space to create a single extra dense point cloud, which enables interesting application scenarios.

\subsection{3D spatio-angular volume extraction}
\label{sec:3DVol}

To obtain a 3D spatio-angular volume, we fix one of the angular dimensions and extract $N$ views over the remaining dimension.
This volume thus consists in a sequence of sub-aperture images, noted $V = \{I_n\}, n = 1 \dots N$.
Here and for the rest of this paper, we assume without loss of generality that we fix $t^*$ and take the  sub-aperture images over $s$ (see Figure~\ref{fig:pipeline}). %($x, y$)

\subsection{Optical flow}

The optical flow method used in this paper was selected for its temporal consistency property~\cite{lang2012practical}.
In our context, this ensures that the different disparity estimates are consistent over the angular dimension.
In addition, we propose to modify the initialization method, as described in the next section.

\subsubsection{Initialization: Coarse-to-fine Patch Matching}
\label{subsec:pm}

In this paper, we propose to use a recent extension of the patch match (PM) method~\cite{barnes2009patchmatch}, the so-called Coarse-to-fine Patch Matching (CPM) technique as initialization~\cite{hu2016efficient}, as it is both more efficient and accurate than the SIFT flow used in~\cite{lang2012practical}.

The well known PM method provides an efficient way to compute sparse matchings between a pair of images.
Given two images $I_1, I_2$, the goal is to find for each patch $p_{1,m}$ in $I_1$ a corresponding patch $p_{2,m} = \mathrm{M}(p_{1,m})$ in $I_2$, with $m=1 \dots M$ where $M$ is the total number of patches in the image.
Note that as we only look for sparse matches, $M$ is much lower than the total number of pixels.
The core idea of PM is to use random search and propagation between neighbors to speed up the matching search.
The matching search itself is conducted as a cost function minimization:

\begin{equation}
\mathrm{M}(p_{1,m}) = \textup{arg}\underset{p_{2,i}}{\textup{min}}~C(p_{1,m}, p_{2,i}), p_{2,i} \in N_{m}
\label{eq:PM}
\end{equation}

\noindent where the cost function $C(\cdot)$ corresponds to the sum of absolute difference (SAD) of the SIFT descriptors, and $N_m$ is a set comprising all patches contained in a search window centered on $p_{1,m}$.
Note that in our context, the sub-aperture images of a light field are rectified, and we can further reduce the complexity of the matching search by limiting the search window to an epipolar line.

%We first describe the CPM correspondence computation between two images $I_1$ and $I_2$.
The CPM method then consists in applying PM on a hierarchical architecture.
A pyramid with $k$ levels is first constructed from the original images with a downsampling factor $\eta$.
This pyramidal decomposition is noted $I_i^l$ with $i=1,2$ and $l = 1 \dots k$.
The PM method is first applied on the $I_1^k$ and $I_2^k$ with a random initialization, and then iteratively on $I_1^l$ and $I_2^l$ with $l = k-1 \dots 1$ using the output of the previous level $l+1$ as initialization.

To initialize our optical flow, we apply the CPM method on consecutive pairs of views $I_n, I_{n+1}$ with $n = 1 \dots N-1$ taken from the volume $V$ built previously, and we note $f_n^{init}$ the flow between these views.

% Due to  higher priority of the accuracy requirement from our following section, a consistency check and outliers handling, as mentioned in~\cite{bao2014fast, hosni2013fast, lu2013patch}, is performed to remove unreliable matches on each level of image pyramid. 

\subsubsection{Efficient spatio-angular filtering: Feature flow}
\label{subsec:tc}
Once sparse matches are obtained as described in previous section, we performed edge-aware filtering on the spatio-angular volume in order to obtain dense consistent correspondences.
%Based on view extraction section, we consider each row or column of images as an angular dimension, which contains sufficient coherent information along continuous  images.
We introduce in this section the feature flow method~\cite{lang2012practical}, an efficient edge-aware filter, used to diffuse sparse matches with coherent information.
%However, in our paper, this filter is performed along angular dimension instead of temporal dimension in original paper.
One of the main advantage of the feature flow is that the global energy minimization operation used in many optical flow approaches is replaced with a local volumetric edge-aware filtering operation.
%, as equation~\eqref{eq:9}, using sparse correspondence data as initialization, which could produce competitive accuracy and significantly improved efficiency.
To properly detect object edges in sub-aperture images and their disparity variations, a domain transform filter~\cite{gastal2011domain} is applied iteratively on the 3 spatio-angular dimensions with a fixed width Gaussian kernel.
The flow $f_{n}$ between views $I_{n}, I_{n+1}$ is obtained from the flow from previous views as $f_{n} = G * f_{n-1}$ where $G$ is the domain transform filter mentioned above and $*$ is a convolution operator. Note that the filtering along the angular dimension follows "disparity paths" provided by the initial sparse flow estimations $f_n^{init}$, with $n = 1 \dots N-1$.

% \begin{equation}
% f_{n} = G * f_{n-1}
% \label{eq:feature_flow}
% \end{equation}

% \noindent where s represents angular dimension, $*$ is a convolution operator, and $G$ is a volumetric filter along spatial and angular direction.
%u represent disparity $\Delta x$ along x dimension, 

%For each iteration of complete filtering, a series of 1D domain transfer filtering iteration is performed along spatial and angular dimension.
%For the filtering  spatial dimension, the filter is applied locally and separately on x, y dimension line by line.
% To filter along the angular dimension, it follows "pseudo motion path" provided by sparse correspondences initialization.
% The initialization of first iteration comes from previous Section~\ref{subsec:pm}, and the output of each iteration is then used as initialization of next iteration.
% Additionally, please note that because initialization correspondences are sparse, there must be some pixels has no previous "pseudo motion path". Thus, the backward initialization optical flow information is used here to create a new path for each homeless pixel.

%Note that as in~\cite{lang2012practical},
To improve the accuracy of the flow estimation, the input sparse flow is weighted with a confidence map, whose weights are computed as the absolute difference between the matching correspondence vectors, thus increasing the contribution of reliable matches.
% Equation~\eqref{eq:feature_flow} can then be rewritten as:
% \begin{equation}
% f_{s+1,t^{*}}(u) = \frac{(G *  (n_{f_{s,t^{*}}}(u)f_{s,t^{*}}(u)))}{(G * n_{f_{s,t^{*}}}(u))} \label{eq:10}
% \end{equation}

% As in~\cite{lang2012practical}, a extended confidence weight $n= \left\{\begin{matrix}
% confid_{x,y} & if\ \ initial_{x,y}\ \ existed\\ 
% 0 & otherwise
% \end{matrix}\right.$ is also introduced to feature flow framework. This confidence weight is assigned to input frame and then both of them is filtered by feature flow to obtain the final results.  In our method, this confidence factor is calculated as absolute difference between matching correspondence vectors, which increase the contribution of reliable sparse matches. 

%\yc{Fig. \ref{} a temporal direction in light field \& image describe motion path}

% \begin{figure}[t]
% 	\begin{center}
% 	\includegraphics[width=17cm,trim={0cm 0cm 2cm 3.5cm},clip]{pipeline_illustrated.pdf}
% 	\end{center}
% 	%\vspace{-0.2cm} 
% 	\caption{Illustration of the different steps of the proposed approach. The method can be apply on any number of rows or columns of the matrix of views in order to obtain more disparity map estimates.}
% \label{fig:pipeline_illustrated}
% %\vspace{-0.2cm}
% \end{figure}

\subsection{From disparity to depth map}
\label{sec:depthmap}
\subsubsection{Single accurate disparity map}

The first approach to obtain the final depth map is to first compute a single disparity map from all estimates, which is then converted into a depth map $Z_{init}$ using the equation from Section~\ref{DPfromLF}.
This technique is used in many state-of-the-art methods, e.g. based on weighted median or image guided filtering, which efficiently removes outliers, and is often followed by a costly global energy minimization technique.

However, thanks to the angular filtering which enforce the consistency between the different disparity map estimates, we can apply a much simpler and faster aggregation step, using median filtering and a one-step variational energy minimization.
Here, the one-step variational energy minimization from the Epicflow~\cite{revaud2015epicflow} method is used to obtain the final depth map $Z_{f}$ from $Z_{init}$:

\begin{equation}
%Z_{f} = \textup{arg} \underset{Z_{i}}{\textup{min}}(C(x,Z_{i})+\lambda \alpha \left \| \bigtriangledown (Z_{i}) \right \|)
Z_{f} = \textup{arg} \underset{Z_{i}}{\textup{min}}(E_{data}(Z_{i}) + \lambda \alpha E_{smooth}(Z_{i}))
\end{equation}

\noindent where $E_{data}$ corresponds to a classical color-constancy data term while $E_{smooth}$ corresponds to a gradient-constancy function with a local smoothness term weight $\alpha = exp(-\kappa \left \| \nabla_{2}Z  \right \|)$~\cite{xu2012motion}, where $\kappa = 5$.

\subsubsection{Extra dense Point Cloud}

%Furthermore,
Thanks to the consistency of the different disparity estimates, we can use a novel process to create the point cloud, where multiple point clouds are created from each disparity map, and then aggregated in the 3D space to obtain the final extra dense point cloud.

\section{Evaluation}
\label{sec:eva}

\begin{figure}[t!]
\centerline{
\includegraphics[width=0.75\linewidth]{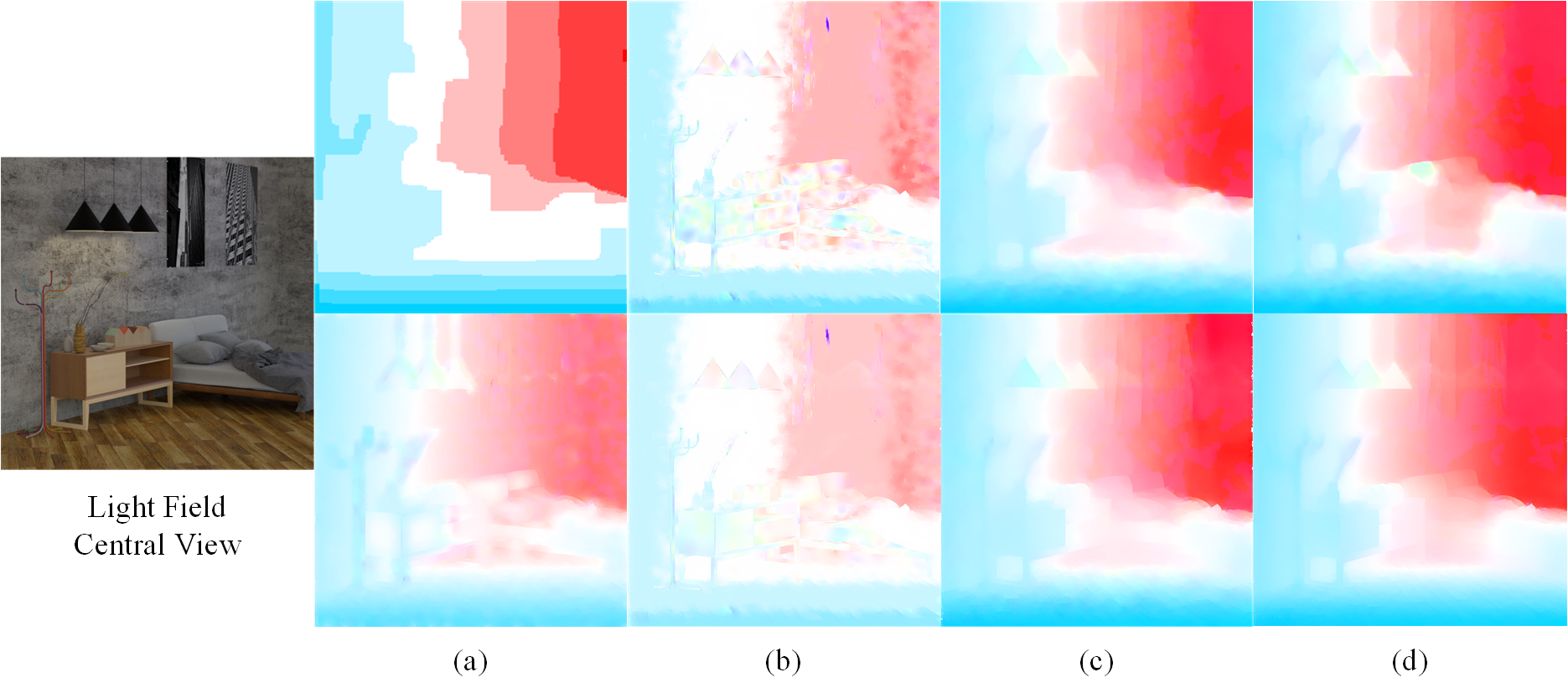}
\vspace{-0.2cm}
}
\caption{Comparison of optical flow obtained with state-of-the-art methods.
%with/without feature flow filtering.
\textbf{Top row} consists of  initialization results with different optical flow methods.
\textbf{Bottom row} is the results of these initializations + feature flow filter.
(a) SIFT Flow~\cite{liu2008sift} (4.9s); (b) EPPM~\cite{bao2014fast} (GPU-based,0.7s); (c) EpicFlow~\cite{revaud2015epicflow} (15s); (d) CPM-Flow~\cite{hu2016efficient} (5.3s) \textbf{(Best viewed in color)}}
\label{fig:initials}
\end{figure}

%In this Section, we verify the accuracy and efficiency of our method mainly based on HCI 4D light field dataset and compare our result with state-of-art light field depth map estimation methods.

In this section, we analyze the results of the proposed approach.
All our experiments were run on an Intel Core i7-6700k 4.0GHz CPU. %are implemented and
We use the feature flow implementation from~\cite{roo2014temporally} and the same parameter setting for all our experiments.
%, and parameters are defined as follow: 
For the CPM method, the level of pyramids $k$ is set to 5, the downsampling factor $\eta$ is set to the 0.5 and the patch size is set to 3x3.
%for feature flow: (all detailed definition of these parameters could be found in domain transform paper~\cite{gastal2011domain}), in spatial dimension, filter factor $\delta_{a}$ is 2000, $\delta_{b}$ is 0.4, and in angular dimension, filter factor $\delta_{as}$ is 5, $\delta_{bs}$ is 0.1.

\textbf{Evaluation of the optical flow.}
We evaluate here the performance of the proposed optical flow approach against state-of-the art methods.
In Figure~\ref{fig:initials}, we show the results of several optical flow initializations in the top row and the results after feature flow filtering in the bottom row.
The volumetric filtering using feature flow along the angular dimension of the light field clearly improves the accuracy of the optical flow from any initialization method, significantly improving consistency and continuity of brightness.
The proposed method using CPM as initialization achieves the best performance in terms of balance between speed and accuracy.

\begin{figure}[t!]
\centerline{
\includegraphics[width=0.8\linewidth]{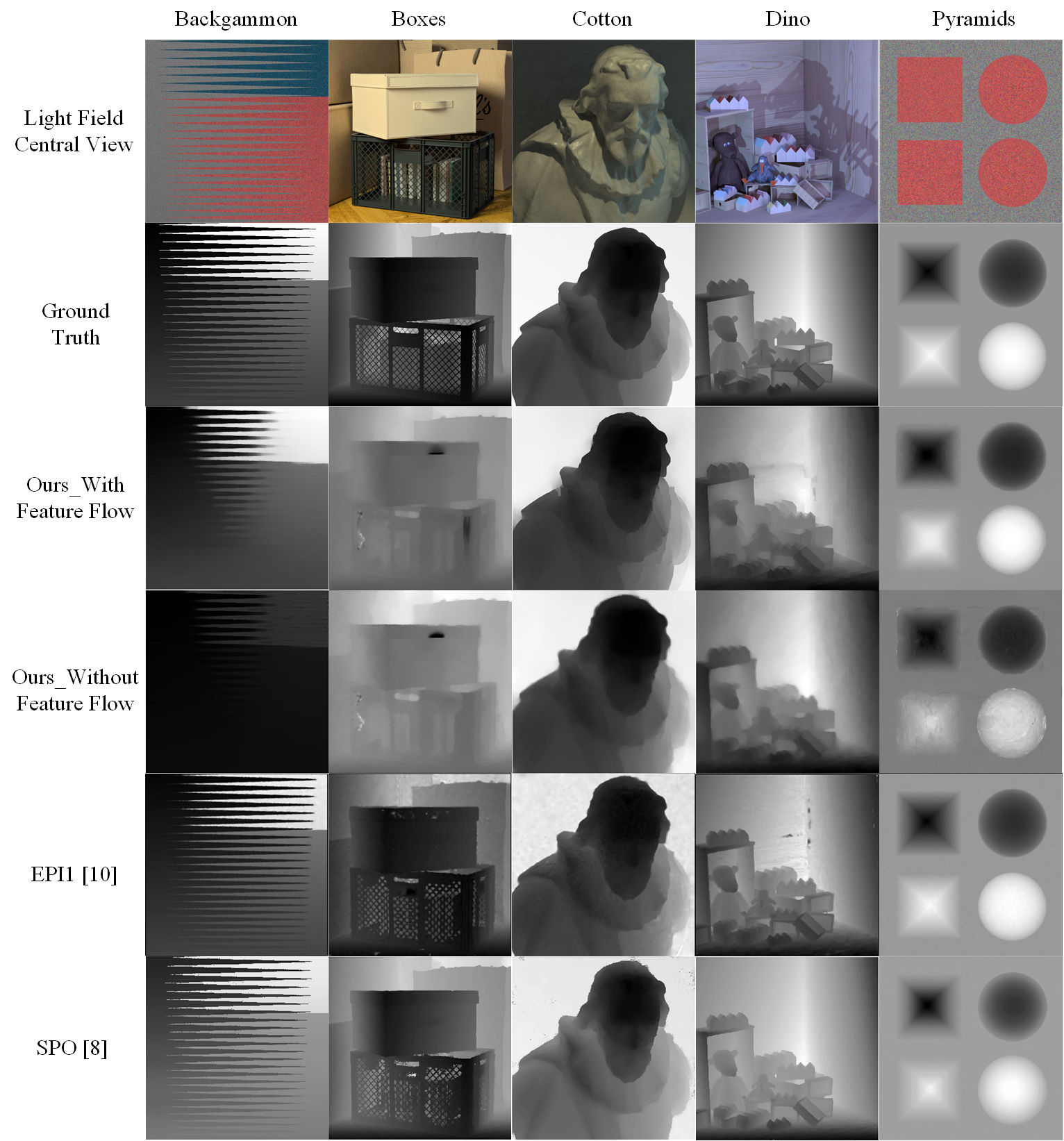}
%\vspace{-2mm}
}
\caption{Depth map comparison on HCI dataset}
\label{fig:depth_map_results}
\end{figure}

The importance of the volumetric filtering is also illustrated in Figure~\ref{fig:depth_map_results}, where we show the final depth map results for several light fields obtained with our method with and without feature flow.
The quality of the depth maps is clearly improved with feature flow for all sequences.

% In order to show the performance of the CPM+feature flow filter, we compare several different state-of-the-art optical flow initialization methods with/without feature flow filter at first. One can see clearly that CPM initialization could produce relatively well smooth flow(compared to EPPM) and clear object edge(compared to SIFT Flow) in Figure~\ref{fig:initials}(top row). Considering from running time aspect, CPM is significantly faster than Epicflow with similar accuracy. Figure~\ref{fig:initials}(bottom row) shows feature flow provides significant improvement in maintenance of brightness continuity(column(a) and column(d)) and ability to reduce area noise(column(c)).

\begin{wrapfigure}{r}{0.5\textwidth}
  \vspace{-20pt}
  \begin{center}
    \includegraphics[width=\linewidth]{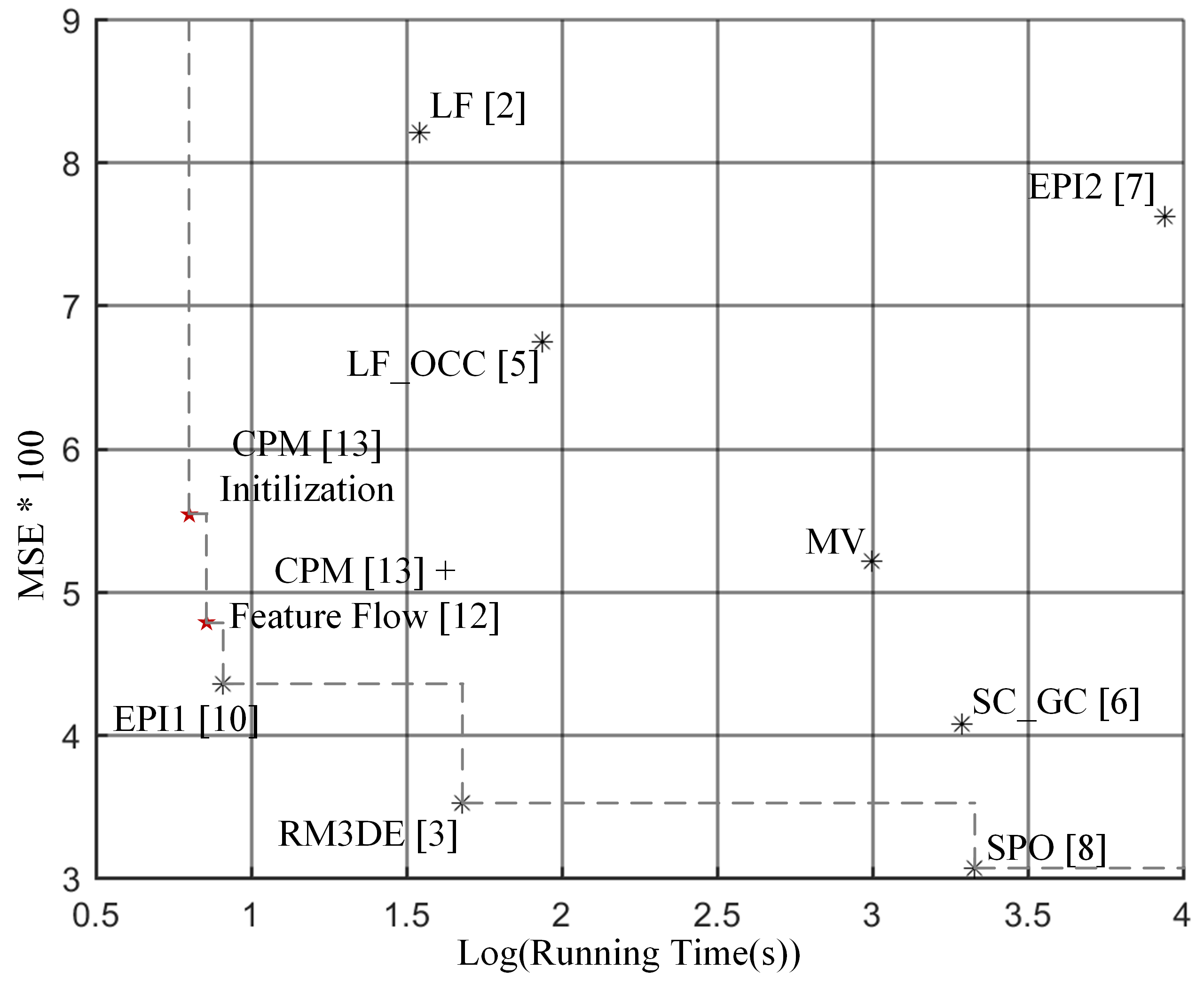}
  \end{center}
\vspace{-20pt}
  \caption{Comparison of our method (red stars) performances against state-of-the-art (blue stars), averaged over all HCI light fields. The results show that we achieve comparable performances to the best state-of-the-art method in terms of balance between speed and accuracy.}
\label{fig:speed_vs_accuracy}
  \vspace{-10pt}
\end{wrapfigure}

\textbf{HCI benchmark performance.} 
We evaluate here the accuracy and efficiency of our proposed method against state-of-the-art light field depth map estimation methods~\cite{Wanner2012a, Johannsen, Zhang2016a, Wang2015a, Jeon2015, Si2017, Neri2015} using the recent HCI 4D light field dataset~\cite{honauer2016benchmark}~\footnote{\url{http://hci-lightfield.iwr.uni-heidelberg.de/}}.
The accuracy of the depth estimation is evaluated using the Mean Square Error (MSE) * 100 and the computational complexity using the running time in seconds.
The results are summed up in the graph of Figure~\ref{fig:speed_vs_accuracy}, showing the average performances over the HCI dataset.
More detailed results will be made available on our web page\footnote{\url{https://v-sense.scss.tcd.ie/?p=842}}. %\footnote{\url{https://v-sense.scss.tcd.ie/}}.
Our method achieves comparable perfomance with the best method of the state-of-the-art in terms of balance between accuracy and speed.

In addition to these objective metrics, we show the depth maps obtained from several light fields in Figure~\ref{fig:depth_map_results} and compare against state-of-the-art methods.
The final comparison shows better performance for edge preservation of objects (see Cotton column) and also smoother results for noisy scenes (see Backgammon and Dino columns).
However, we notice that the proposed local filtering method, which allows considerable speed up, sometime produces less smooth results than a global solution (see background of Dino and Boxes column).
The feature flow filter also heavily depends on the quality of the optical flow initialization.
If the optical flow method is unable to provide accurate correspondence, it can not be corrected by the filter (see for example the Boxes column).
% Admittedly, our method fails to find global optimal solution due to local optimization(see background of Dino and boxes column). However, it may be worthy since this proposed method has significant advantage in processing time shown in Figure~\ref{fig:speed_vs_accuracy}.

Note that at the time of writing, some recent results were added in the HCI benchmark without any attached publications. As we can not fully understand and exploit comparisons with methods which are not described, we choose in this paper to compare our results only against published work.

% On HCI dataset, the Mean Square Error(MSE)*100 and Running Time(log10) are used to evaluate all these methods on several light field sequences. The final comparison result is shown in Figure~\ref{fig:speed_vs_accuracy}. The results shows that our method outperforms all other methods in running time in several HCI light field image sequences. Besides, compared to most accurate methods, we also performs closely well accuracy and require less computation cost. The full depth map result could be observed in Figure~\ref{fig:depth_map_results}.  Readers could obtain complete description from their paper~\cite{honauer2016benchmark} and website(\url{}) if interested.

\begin{wrapfigure}{r}{0.5\textwidth}
  \vspace{-20pt}
  \begin{center}
    \includegraphics[width=\linewidth]{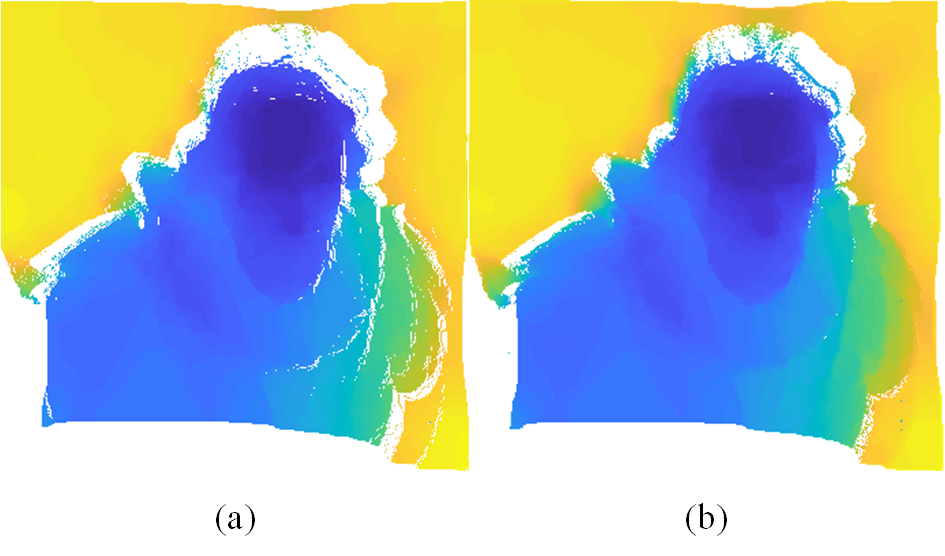}
  \end{center}
\vspace{-20pt}
  \caption{(a) Single point clouds (about 260k points) v.s. (b) Extra dense point clouds (more than 21 million points). \textbf{(Best viewed in color)}}
\label{fig:multi_single}
%   \vspace{-0pt}
\end{wrapfigure}

\textbf{Extra dense 3D Point Cloud.}
As mentioned in Section~\ref{sec:depthmap}, a unique feature of our method compared to the state-of-the-art approaches is that we can produce several consistent depth maps which can then be integrated in the 3D space in order to produce extra dense point clouds.
Examples of such results are shown in Figure~\ref{fig:multi_single} in comparison with point clouds obtained with a single depth map.

% one disparity map for each image in a 4D light field, which could be integrated into one super dense 3D reconstruction(point cloud). Figure~\ref{fig:multi_single} demonstrates a comparison between this extra dense point cloud and common point cloud from single central view disparity map. This provides many possibilities of potential applications(e.g. virtual reality in light field), which may require higher accuracy and continuity, to be explored in the future.

\section{Conclusion}
\label{sec:conclusion}

In this paper, we introduced a novel optical flow-based method to estimate depth maps from light fields.
We showed that by extracting a 3D volume consisting of a sequence of views from the 4D light field, and applying a temporally consistent optical flow on this spatio-angular volume, we were able to obtain high-quality depth maps with a reduced complexity.
Comparison with state-of-the-art methods on the HCI benchmark showed that we are competitive with the best method in terms of balance between accuracy and speed.
Furthermore, thanks to the enforced consistency of the disparity map estimates along the angular dimension, we are able to produce point clouds with a much higher density than in any state-of-the-art method.
However, we note that our local filtering sometimes yield inferior results compared to global optimizations.
In future work, we plan to investigate different spatio-angular filtering methods in order to improve the accuracy while keeping a faster running time.
Furthermore, we intend to apply our method on dense light fields captured with lenslet cameras such as Lytro~\cite{Lytro} to perform 3D reconstruction of real world scenes.

% \yc{To summarize the paper and give some possible limitations and future work}
% We have introduced a rapid temporal filter with reliable matching initialization for 4D light field depth map estimation. We estimate correspondence matching firstly with patch match method and a top-down hierarchical structure. All above work are  preformed in vertical and horizon direction of light field image matrix and then globally integrated with a variational refinement. Compared to state-of-art results, our method demonstrates competitive accuracy and significantly faster runtime and lower computation cost, even though GPU parallelism is not used for acceleration. 
% Our method still has some improvement space. A primary problem for our temporal filter is that poor performance could happen when object boundary can't be represented by image edge. Our local optimization method may lead discontinuity when initial matches are too sparse. Besides, it also could fail when facing very noisy input image sequence. One potential solution is to applied a de-noise algorithm as pre-process, which however, may cause high computation cost. Future work could focus on these two aspects. Despite these limitations, our method still is an competitive tool with accuracy and efficiency for depth map estimation method of light field images which may be used in various related practical applications.

% \printbibliography

% \bibliographystyle{alphaurl}
% \bibliographystyle{apalike}
% \bibliography{ref}
\bibliographystyle{plain}
\bibliography{rtf}

\end{document}